\documentclass[letterpaper]{article} % DO NOT CHANGE THIS
\usepackage{aaai24}  % DO NOT CHANGE THIS
\usepackage{times}  % DO NOT CHANGE THIS
\usepackage{helvet}  % DO NOT CHANGE THIS
\usepackage{courier}  % DO NOT CHANGE THIS
\usepackage[hyphens]{url}  % DO NOT CHANGE THIS
\usepackage{graphicx} % DO NOT CHANGE THIS
\usepackage{natbib}  % DO NOT CHANGE THIS AND DO NOT ADD ANY OPTIONS TO IT
\usepackage{caption} % DO NOT CHANGE THIS AND DO NOT ADD ANY OPTIONS TO IT
\usepackage{algorithm}
\usepackage{algorithmic}
\usepackage{mathtools}
\usepackage[nolist]{acronym}
\usepackage{tikz}
\usetikzlibrary{calc}
\usepackage{amsmath}
\usepackage[inline]{enumitem}

\usepackage{newfloat}
\usepackage{listings}
\DeclareCaptionStyle{ruled}{labelfont=normalfont,labelsep=colon,strut=off} % DO NOT CHANGE THIS
\floatstyle{ruled}
\newfloat{listing}{tb}{lst}{}
\floatname{listing}{Listing}
\title{SAT-Based Algorithms for Regular Graph Pattern Matching\footnote{This is an extended version of a paper accepted for publication in the main technical track of the AAAI 2024 conference.}}
\author{
    Miguel Terra-Neves\textsuperscript{\rm 1},
    José Amaral\textsuperscript{\rm 1},
    Alexandre Lemos\textsuperscript{\rm 1},\\
    Rui Quintino\textsuperscript{\rm 1},
    Pedro Resende\textsuperscript{\rm 2},
    Antonio Alegria\textsuperscript{\rm 1}
}
\affiliations{
    \textsuperscript{\rm 1}OutSystems\\
    \textsuperscript{\rm 2}Zharta\\
    miguel.neves@outsystems.com,
    jose.francisco.amaral@outsystems.com,
    alexandre.lemos@outsystems.com,\\
    rui.quintino@outsystems.com,
    pt.resende@icloud.com,
    antonio.alegria@outsystems.com
}

\newtheorem{example}{Example}
\newtheorem{definition}{Definition}
\newtheorem{proposition}{Proposition}
\newtheorem{proof}{Proof}

\begin{document}

\maketitle

\begin{abstract}
Graph matching is a fundamental problem in pattern recognition, with many applications such as software analysis and computational biology.
One well-known type of graph matching problem is graph isomorphism, which consists of deciding if two graphs are identical.
Despite its usefulness, the properties that one may check using graph isomorphism are rather limited, since it only allows strict equality checks between two graphs.
For example, it does not allow one to check complex structural properties such as if the target graph is an arbitrary length sequence followed by an arbitrary size loop.

We propose a generalization of graph isomorphism that allows one to check such properties through a declarative specification.
This specification is given in the form of a Regular Graph Pattern (ReGaP), a special type of graph, inspired by regular expressions, that may contain wildcard nodes that represent arbitrary structures such as variable-sized sequences or subgraphs.
We propose a SAT-based algorithm for checking if a target graph matches a given ReGaP.
We also propose a preprocessing technique for improving the performance of the algorithm and evaluate it through an extensive experimental evaluation on benchmarks from the CodeSearchNet dataset.
\end{abstract}

\begin{acronym}
    \acro{ReGaP}{Regular Graph Pattern}
    \acro{CNF}{Conjunctive Normal Form}
    \acro{SAT}{Boolean Satisfiability}
    \acro{CFG}{Control Flow Graph}
\end{acronym}

\section{Introduction}
\label{sec:intro}

Pattern recognition is an important research area~\cite{DBLP:journals/ijprai/FoggiaPV14} due to its numerous applications ranging from detecting bad code patterns~\cite{DBLP:series/lndect/PiotrowskiM20} and software analysis~\cite{DBLP:conf/csiirw/ParkRMS10,Singh2021,DBLP:conf/kbse/ZouBXX20} in general, to computational biology~\cite{DBLP:conf/iciap/CarlettiFV13,DBLP:journals/bioinformatics/ZaslavskiyBV09}.
One fundamental problem in pattern recognition is graph matching~\cite{DBLP:journals/paa/LiviR13}.
Two common approaches are: (i) graph isomorphism~\cite{DBLP:conf/iciap/CordellaFSV99,DBLP:conf/icpr/DahmBCG12,DBLP:journals/jacm/Ullmann76,DBLP:journals/jea/Ullmann10,DBLP:journals/mscs/LarrosaV02,DBLP:journals/jea/Ullmann10,DBLP:journals/constraints/ZampelliDS10} and (ii) approximated graph matching~\cite{DBLP:journals/prl/Bunke97,DBLP:journals/jcamd/RaymondW02a,DBLP:journals/tsmc/SanfeliuF83}.
The first consists of deciding if two graphs are identical, which can be too strict for some applications~\cite{DBLP:journals/prl/Auwatanamongkol07,DBLP:journals/ijprai/ConteFSV04}.
Approximated graph matching algorithms are less strict and normally employ some sort of distance metric to evaluate the graphs.
Despite their usefulness, these approaches suffer from limitations.
Neither of these allows one to check if a graph satisfies some specific complex structural properties, such as, for example, if some target graph contains two nested cycles since, given another similar reference graph that satisfies that property, one can increase their distance arbitrarily by adding extra nodes to, for example, the inner cycle.

Alternatively, regular-path queries~\cite{DBLP:conf/sigmod/CruzMW87} allow one to specify paths between nodes through a regular expression.
Different formalisms for this type of query exist in the literature~\cite{DBLP:conf/sigmod/AnglesABBFGLPPS18,DBLP:journals/fcsc/FanLMTW12,DBLP:journals/mst/ReutterRV17,DBLP:conf/dexa/WangWXZZ20,DBLP:conf/semweb/ZhangFWRW16,DBLP:conf/icdt/LibkinMV13}.
These are very expressive and useful, but do not allow one to check if a graph contains some complex subgraph structure.
For example, it is not possible to specify a sequence of arbitrary nested loops with no external connections.
Therefore, we propose \acp{ReGaP} as a generalization of graph isomorphism.
The goal is to be able to define complex structural properties through a declarative specification in the form of a special graph.
The proposed specification also borrows inspiration from regular expressions.
This graph may contain special nodes, referred to as wildcards, representing arbitrary structures such as variable-sized sequences or subgraphs.
These wildcards enable one to define compact representations of infinite sets of graphs.

The main contributions of this paper are three-fold: 
(i) a generalization of graph isomorphism matching in the form of \ac{ReGaP} matching;
(ii) a novel \ac{SAT} encoding for the \ac{ReGaP} matching problem; and
(iii) a graph simplification technique for improving the performance of the \ac{SAT} solver.
The proposed solution is evaluated using control-flow graphs extracted from the Python code snippets in the CodeSearchNet dataset~\cite{DBLP:journals/corr/abs-1909-09436}.
The \acp{ReGaP} replicate the kind of bad code
patterns that are integrated in the AI Mentor Studio~\cite{aimentorstudio} code analysis engine for the OutSystems visual programming language.
Note that, although the evaluation focuses on a specific use case, the concept and algorithm are generic, and thus may be applied in other contexts.

\section{Background}
\label{sec:background}

In this section, we introduce the necessary background.
We start with a brief introduction to graph isomorphism in Section~\ref{sec:isomorphism}, followed by an explanation of \ac{SAT} in Section~\ref{sec:sat}.

\subsection{Graph Isomorphism}
\label{sec:isomorphism}

Consider two graphs $G_1 = (V_1, E_1)$ and $G_2 = (V_2, E_2)$. $G_1$ and $G_2$ are isomorphic if and only if there exists a bijective mapping $f: V_1 \leftrightarrow V_2$ between the nodes of $V_1$ and $V_2$ such that  for all $(u,v) \in E_1$, $(f(u), f(v)) \in E_2$ and vice-versa.

In an attributed graph $G = (V, E)$, each node has $m$ attributes, denoted as $A_V = \{a_V^1, \ldots, a_V^m\}$.
A node $v \in V$ is associated with an attribute vector $A_V(v) = \left[a_V^1(v), \ldots, a_V^m(v) \right]$, where $a_V^i(v)$ is the value of attribute $a_V^i$ for node $v$.
Similarly, each edge has $n$ attributes, denoted as $A_E = \{a_E^1, \ldots, a_E^n\}$, and $A_E((u, v)) = \left[ a_E^1((u, v)), \dots, a_E^n((u, v)) \right]$ is the attribute vector for edge $(u, v)$.
The definition of graph isomorphism must ensure consistency between all the attributes of a node/edge of $G_1$ and the respective equivalent in $G_2$, i.e. for all $u \in V_1$, $A_V(u) = A_V(f(u))$, and for all $(u, v) \in E_1$, $A_E((u,v)) = A_E((f(u), f(v)))$.

\subsection{Boolean Satisfiability}
\label{sec:sat}

Let $X$ be a set of Boolean variables.
A literal $l$ is either a variable $x \in X$ or its negation $\overline{x}$.
A clause $c$ is a disjunction of literals $(l_1 \vee \dots \vee l_k)$.
A propositional logic formula $F$ in \ac{CNF} is a conjunction of clauses $c_1 \wedge \dots \wedge c_n$.
A complete assignment $\alpha: X \rightarrow \lbrace 0, 1 \rbrace$ is a function that assigns a Boolean value to each variable in $X$.
A literal $x$ ($\overline{x}$) is satisfied by $\alpha$ if and only if $\alpha(x) = 1$ ($\alpha(x) = 0$).
A clause $c$ is satisfied by $\alpha$ if and only if at least one of its literals is satisfied.
A \ac{CNF} formula $F$ is satisfied by $\alpha$ if and only if all of its clauses are satisfied.
Given a \ac{CNF} formula $F$, the \ac{SAT} problem consists of deciding if there exists $\alpha$ which satisfies $F$.
If so, then $F$ is satisfiable and $\alpha$ is a model of $F$.
Otherwise, $F$ is unsatisfiable.
Nowadays, most \ac{SAT} solvers implement the conflict-driven clause learning algorithm~\cite{glucose,DBLP:conf/ijcai/AudemardS09,BiereFleuryHeisinger-SAT-Competition-2021-solvers,DBLP:conf/sat/LiangOMTLG18,DBLP:conf/iccad/SilvaS96,SLIME}.
Further details can be found in the literature~\cite{handbook}.

\section{Problem Definition}
\label{sec:problem-def}

The \ac{ReGaP} matching problem consists of determining if a given \ac{ReGaP} $P = (V_P, E_P)$ matches some graph $G = (V, E)$.
We assume that $G$ is non-attributed for now.
A \ac{ReGaP} is a graph such that some of the nodes in $V_P$ may be of a special type referred to as wildcard.
We consider four wildcard types, inspired on regular expressions:
\begin{itemize}
    \item \textbf{any-1+-sequence} (\textbf{any-0+-sequence}).
    Represents a directed path $v^1, \dots, v^k$ of 1 (0) or more nodes such that, for each $i \in \lbrace 2..k \rbrace$, the only edge in $E$ towards $v^i$ is $(v^{i-1}, v^i)$.
    We use $W_P^{S+} \subseteq V_P$ ($W_P^{S*} \subseteq V_P$) to denote the set of all any-1+-sequence (any-0+-sequence) wildcards in $V_P$.
    \item \textbf{any-1+-subgraph} (\textbf{any-0+-subgraph}).
    Represents a subgraph of 1 (0) or more nodes.
    We use $W_P^{G+} \subseteq V_P$ ($W_P^{G*} \subseteq V_P$) to denote the set of all any-1+-subgraph (any-0+-subgraph) wildcards in $V_P$.
\end{itemize}

\begin{figure}[t]
    \centering
    \begin{tikzpicture}
      \node[circle, draw] (v1) {$v_1$};
      \node[circle, right of=v1, draw, xshift=0.8em] (v2) {$v_2$};
      \node[circle, right of=v2, draw, xshift=0.8em] (v3) {$v_3$};
      \node[circle, right of=v3, draw, xshift=0.8em] (v4)  {$v_4$};
      \node[circle, right of=v4, draw, xshift=0.8em] (v5) {$v_5$};
      \node[circle, below of=v3, draw, xshift=0.8em] (v6) {$v_6$};
      \node[circle, below of=v5, draw, xshift=-0.8em] (v7) {$v_7$};
      
      \draw[->] (v1) -- (v2);
      \draw[->] (v2) -- (v3);
      \draw[->] (v3) -- (v4);
      \draw[->] (v4.north) |- ++(0.3, 0.3) -| (v5.north);
      \draw[<-] (v4) -- (v5);
      \draw[->] (v4) -- (v6);
      \draw[->] (v4) -- (v7);  
    
      \node[circle, below of=v1, draw, yshift=-4em, xshift=1.5em] (A) {A};
      \node[right of=A, draw, xshift=0.8em] (S) {$S+$};
      \node[circle, right of=S, draw, xshift=0.8em] (B) {B};
      \node[circle, right of=B, draw, xshift=0.8em] (C) {C};
      \node[draw, below of=B] (G) {$G+$};
      
      \draw[->] (A) -- (S);
      \draw[->] (S) -- (B);
      \draw[->] (B.north) |- ++(0.3, 0.3) -| (C.north);
      \draw[->] (B) -- (G);
      \draw[<-] (B) -- (C);
    \end{tikzpicture}
    \caption{An example of a graph (top) and a \ac{ReGaP} (bottom) that matches that graph. $S+$ represents an any-1+-sequence wildcard and $G+$ an any-1+-subgraph.}
    \label{fig:running_example}
\end{figure}
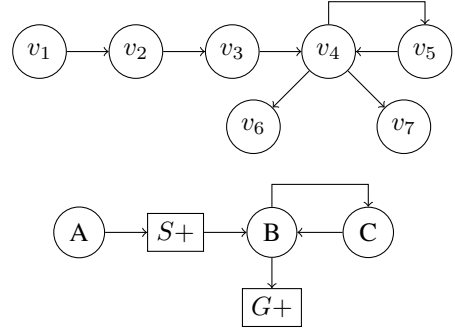

\begin{example}
    Figure~\ref{fig:running_example} shows an example of a graph $G$ and \ac{ReGaP} $P$ that matches $G$.
    $P$ contains an any-1+-sequence wildcard $S+$ and an any-1+-subgraph wildcard $G+$.
\end{example}

We use $W_P^+$ ($W_P^*$) to denote the set of all any-1+ (any-0+) wildcards, i.e. $W_P^+ = W_P^{S+} \cup W_P^{G+}$ ($W_P^* = W_P^{S*} \cup W_P^{G*}$), $W_P^S$ ($W_P^G$) to denote the set of all sequence (subgraph) wildcards, i.e. $W_P^S = W_P^{S+} \cup W_P^{S*}$ ($W_P^G = W_P^{G+} \cup W_P^{G*}$), and $W_P$ to denote the set of all wildcards, i.e. $W_P = W_P^S \cup W_P^G$.

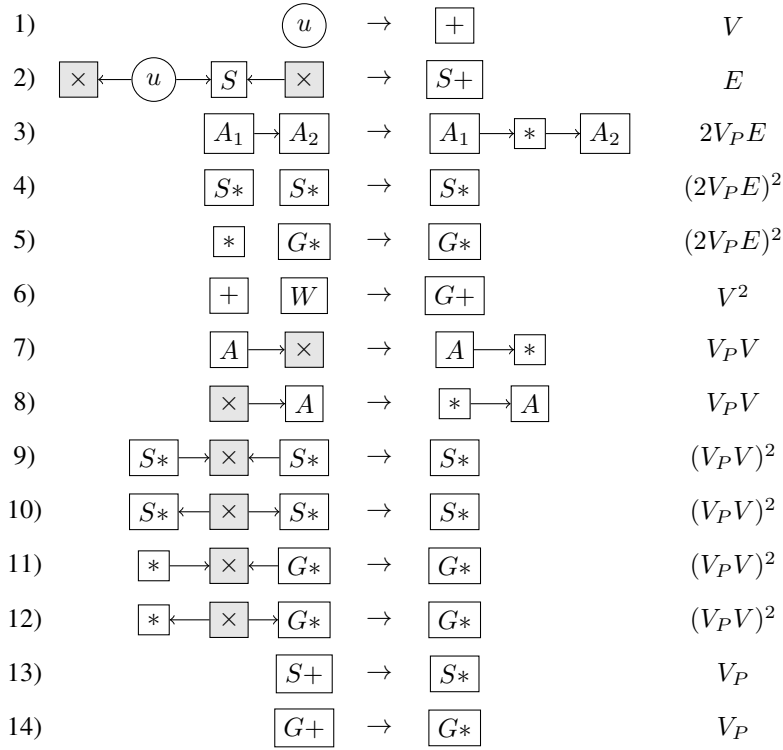
\begin{figure*}[t]
    \centering
    \begin{tikzpicture}
        % Prefix rule 0
        \node[draw, circle](rule0-u){$u$};
        \node[right of=rule0-u](rule0-arrow){$\rightarrow$};
        \node[draw, right of=rule0-arrow](rule0-1p){$+$};
        % Prefix rule 2
        \node[below of=rule0-arrow, yshift=0.8em](rule2-arrow){$\rightarrow$};
        \node[draw, left of=rule2-arrow, fill=black!10](rule2-anti){$\times$};
        \node[draw, left of=rule2-anti](rule2-s){$S$};
        \node[draw, circle, left of=rule2-s](rule2-u){$u$};
        \node[draw, left of=rule2-u, fill=black!10](rule2-anti-u){$\times$};
        \path[->, draw] (rule2-anti) edge node {} (rule2-s);
        \path[->, draw] (rule2-u) edge node {} (rule2-s);
        \path[->, draw] (rule2-u) edge node {} (rule2-anti-u);
        \node[draw, right of=rule2-arrow](rule2-0p-seq){$S+$};
        % Prefix rule 1
        \node[below of=rule2-arrow, yshift=0.8em](rule1-arrow){$\rightarrow$};
        \node[draw, left of=rule1-arrow](rule1-v){$A_2$};
        \node[draw, left of=rule1-v](rule1-u){$A_1$};
        \path[->, draw] (rule1-u) edge node {} (rule1-v);
        \node[draw, right of=rule1-arrow](rule1-new-u){$A_1$};
        \node[draw, right of=rule1-new-u](rule1-new-0p){$*$};
        \path[->, draw] (rule1-new-u) edge node {} (rule1-new-0p);
        \node[draw, right of=rule1-new-0p](rule1-new-v){$A_2$};
        \path[->, draw] (rule1-new-0p) edge node {} (rule1-new-v);
        % Prefix rule 4
        \node[below of=rule1-arrow, yshift=0.8em](rule4-arrow){$\rightarrow$};
        \node[draw, left of=rule4-arrow](rule4-0p-seq1){$S*$};
        \node[draw, left of=rule4-0p-seq1](rule4-0p-seq2){$S*$};
        \node[draw, right of=rule4-arrow](rule4-new-0p-seq){$S*$};
        % Prefix rule 5
        \node[below of=rule4-arrow, yshift=0.8em](rule5-arrow){$\rightarrow$};
        \node[draw, left of=rule5-arrow](rule5-0p-g){$G*$};
        \node[draw, left of=rule5-0p-g](rule5-0p){$*$};
        \node[draw, right of=rule5-arrow](rule4-new-0p-g){$G*$};
        % Prefix rule 6
        \node[below of=rule5-arrow, yshift=0.8em](rule6-arrow){$\rightarrow$};
        \node[draw, left of=rule6-arrow](rule6-w){$W$};
        \node[draw, left of=rule6-w](rule6-1p){$+$};
        \node[draw, right of=rule6-arrow](rule6-1p-g){$G+$};
        % Suffix rule 3
        \node[below of=rule6-arrow, yshift=0.8em](s-rule3-arrow){$\rightarrow$};
        \node[draw, left of=s-rule3-arrow, fill=black!10](s-rule3-anti){$\times$};
        \node[draw, left of=s-rule3-anti](s-rule3-any){$A$};
        \path[->, draw] (s-rule3-any) edge node {} (s-rule3-anti);
        \node[draw, right of=s-rule3-arrow](s-rule3-new-any){$A$};
        \node[draw, right of=s-rule3-new-any](s-rule3-0p){$*$};
        \path[->, draw] (s-rule3-new-any) edge node {} (s-rule3-0p);
        % Suffix rule 4
        \node[below of=s-rule3-arrow, yshift=0.8em](s-rule4-arrow){$\rightarrow$};
        \node[draw, left of=s-rule4-arrow](s-rule4-any){$A$};
        \node[draw, left of=s-rule4-any, fill=black!10](s-rule4-anti){$\times$};
        \path[->, draw] (s-rule4-anti) edge node {} (s-rule4-any);
        \node[draw, right of=s-rule4-arrow](s-rule4-0p){$*$};
        \node[draw, right of=s-rule4-0p](s-rule4-new-any){$A$};
        \path[->, draw] (s-rule4-0p) edge node {} (s-rule4-new-any);
        % Suffix rule 5
        \node[below of=s-rule4-arrow, yshift=0.8em](s-rule5-arrow){$\rightarrow$};
        \node[draw, left of=s-rule5-arrow](s-rule5-0p-seq1){$S*$};
        \node[draw, left of=s-rule5-0p-seq1, fill=black!10](s-rule5-anti){$\times$};
        \node[draw, left of=s-rule5-anti](s-rule5-0p-seq2){$S*$};
        \path[->, draw] (s-rule5-0p-seq1) edge node {} (s-rule5-anti);
        \path[->, draw] (s-rule5-0p-seq2) edge node {} (s-rule5-anti);
        \node[draw, right of=s-rule5-arrow](s-rule5-new-0p-seq){$S*$};
        % Suffix rule 6
        \node[below of=s-rule5-arrow, yshift=0.8em](s-rule6-arrow){$\rightarrow$};
        \node[draw, left of=s-rule6-arrow](s-rule6-0p-seq1){$S*$};
        \node[draw, left of=s-rule6-0p-seq1, fill=black!10](s-rule6-anti){$\times$};
        \node[draw, left of=s-rule6-anti](s-rule6-0p-seq2){$S*$};
        \path[->, draw] (s-rule6-anti) edge node {} (s-rule6-0p-seq1);
        \path[->, draw] (s-rule6-anti) edge node {} (s-rule6-0p-seq2);
        \node[draw, right of=s-rule6-arrow](s-rule6-new-0p-seq){$S*$};
        % Suffix rule 7
        \node[below of=s-rule6-arrow, yshift=0.8em](s-rule7-arrow){$\rightarrow$};
        \node[draw, left of=s-rule7-arrow](s-rule7-0p-g){$G*$};
        \node[draw, left of=s-rule7-0p-g, fill=black!10](s-rule7-anti){$\times$};
        \node[draw, left of=s-rule7-anti](s-rule7-any){$*$};
        \path[->, draw] (s-rule7-0p-g) edge node {} (s-rule7-anti);
        \path[->, draw] (s-rule7-any) edge node {} (s-rule7-anti);
        \node[draw, right of=s-rule7-arrow](s-rule7-new-0p-g){$G*$};
        % Suffix rule 8
        \node[below of=s-rule7-arrow, yshift=0.8em](s-rule8-arrow){$\rightarrow$};
        \node[draw, left of=s-rule8-arrow](s-rule8-0p-g){$G*$};
        \node[draw, left of=s-rule8-0p-g, fill=black!10](s-rule8-anti){$\times$};
        \node[draw, left of=s-rule8-anti](s-rule8-any){$*$};
        \path[->, draw] (s-rule8-anti) edge node {} (s-rule8-0p-g);
        \path[->, draw] (s-rule8-anti) edge node {} (s-rule8-any);
        \node[draw, right of=s-rule8-arrow](s-rule8-new-0p-g){$G*$};
        % Suffix rule 1
        \node[below of=s-rule8-arrow, yshift=0.8em](s-rule1-arrow){$\rightarrow$};
        \node[draw, left of=s-rule1-arrow](s-rule1-1p-seq){$S+$};
        \node[draw, right of=s-rule1-arrow](s-rule1-0p-seq){$S*$};
        % Suffix rule 2
        \node[below of=s-rule1-arrow, yshift=0.8em](s-rule2-arrow){$\rightarrow$};
        \node[draw, left of=s-rule2-arrow](s-rule1-1p-g){$G+$};
        \node[draw, right of=s-rule2-arrow](s-rule1-0p-g){$G*$};
        % Rule numbers
        \node[left of=rule2-anti-u, xshift=0.8em](rule2-num){2)};
        \node[above of=rule2-num, yshift=-0.8em](rule0-num){1)};
        \node[below of=rule2-num, yshift=0.8em](rule1-num){3)};
        \node[below of=rule1-num, yshift=0.8em](rule3-num){4)};
        \node[below of=rule3-num, yshift=0.8em](rule4-num){5)};
        \node[below of=rule4-num, yshift=0.8em](rule5-num){6)};
        \node[below of=rule5-num, yshift=0.8em](s-rule1-num){7)};
        \node[below of=s-rule1-num, yshift=0.8em](s-rule2-num){8)};
        \node[below of=s-rule2-num, yshift=0.8em](s-rule3-num){9)};
        \node[below of=s-rule3-num, yshift=0.8em](s-rule4-num){10)};
        \node[below of=s-rule4-num, yshift=0.8em](s-rule5-num){11)};
        \node[below of=s-rule5-num, yshift=0.8em](s-rule6-num){12)};
        \node[below of=s-rule6-num, yshift=0.8em](s-rule7-num){13)};
        \node[below of=s-rule7-num, yshift=0.8em](s-rule8-num){14)};
        % Worst case rule applications
        \node[right of=rule1-new-v, xshift=2em](rule1-worst){$2 V_P E$};
        \node[above of=rule1-worst, yshift=-0.8em](rule2-worst){$E$};
        \node[above of=rule2-worst, yshift=-0.8em](rule0-worst){$V$};
        \node[below of=rule1-worst, yshift=0.8em](rule3-worst){$(2 V_P E)^2$};
        \node[below of=rule3-worst, yshift=0.8em](rule4-worst){$(2 V_P E)^2$};
        \node[below of=rule4-worst, yshift=0.8em](rule5-worst){$V^2$};
        \node[below of=rule5-worst, yshift=0.8em](s-rule1-worst){$V_P V$};
        \node[below of=s-rule1-worst, yshift=0.8em](s-rule2-worst){$V_P V$};
        \node[below of=s-rule2-worst, yshift=0.8em](s-rule3-worst){$(V_P V)^2$};
        \node[below of=s-rule3-worst, yshift=0.8em](s-rule4-worst){$(V_P V)^2$};
        \node[below of=s-rule4-worst, yshift=0.8em](s-rule5-worst){$(V_P V)^2$};
        \node[below of=s-rule5-worst, yshift=0.8em](s-rule6-worst){$(V_P V)^2$};
        \node[below of=s-rule6-worst, yshift=0.8em](s-rule7-worst){$V_P$};
        \node[below of=s-rule7-worst, yshift=0.8em](s-rule8-worst){$V_P$};
    \end{tikzpicture}
    \caption{Rules for generalizing a graph $G$. On the right is the worst-case number of applications of each rule.}
    \label{fig:gen-rules}
\end{figure*}

The definition of matching between a \ac{ReGaP} and a graph $G = (V, E)$ relies on a set of generalization rules depicted in Figure~\ref{fig:gen-rules}, which transform $G$ into a generalized version $G'$, i.e. $G'$ is a \ac{ReGaP} that matches $G$.
For example, rule~1 replaces a non-wildcard $u \in V$ by an any-1+ wildcard, represented by the $+$ node.
$W$ represents any wildcard type, while $A$ represents any node.
The rules must be applied in order\footnote{Note that some rules are actually commutative, such as 7 and 8. However, others do need to follow the defined order, otherwise the definition would allow matches that do not make sense.
For simplicity, a strict order is considered instead of a partial one.} (e.g. rule~1 cannot be applied after an instance of rule~2).
By default, no constraint is imposed on the subgraph in the left-hand side of a rule and the respective connections to other nodes in $V$.
The special anti-node $\times$ is used to specify such constraints.
For example, the second anti-node in rule~2 dictates that no edge $(u', S) \in E$ may exist such that $u' \neq u$.
Such anti-nodes prevent non-directed paths from being generalized into an any-1+-sequence.
For example, consider a generalized graph of the form $u \rightarrow S+ \leftarrow v$.
One cannot apply rule 2 on the edge $(u, S+)$ due to the aforementioned anti-node and the existence of the edge $(v, S+)$.

\begin{definition}\label{def:regap-match}
Let $P = (V_P, E_P)$ be a \ac{ReGaP} and $G$ a graph.
$P$ is said to match $G$ if and only if there exists a sequence of generalization rules transforming $G$ into $G' = (V', E')$ such that there exists a bijective mapping $f: V_P \leftrightarrow V'$ that satisfies the following conditions:
\begin{enumerate}
    \item for all $(u, v) \in E_P$, $(f(u), f(v)) \in E'$ and vice-versa.
    \item for all $w \in W_P$, $f(w)$ is a wildcard of the same type.
\end{enumerate}
\end{definition}

\begin{example}
Consider the example from Figure~\ref{fig:running_example}.
By applying rules 1 and 2 to replace $v_2$ and $v_3$ with $S+$ and rules 1 and 6 to replace $v_6$ and $v_7$ with $G+$ one obtains a generalized graph that satisfies the conditions of Definition~\ref{def:regap-match}.
\end{example}

Regarding time complexity, consider a non-deterministic algorithm that guesses the sequence of rule applications and the certificate of isomorphism proving that the resulting generalized graph does match $P$.
In the worst case, rule 1 is applied to each node in $V$ and rule 2 to each edge in $E$.
Similarly, rule 3 is applied at most $2 V_P E$ times: 2 for both any-0+ wildcard types, and $V_P$ because $V_P$ may contain only any-0+ wildcards.
The worst case for each rule is shown in Figure~\ref{fig:gen-rules}.
The total worst case is polynomial, and thus \ac{ReGaP} matching is in NP.

Graph isomorphism is a special case of \ac{ReGaP} matching.
The introduction of wildcards enables the compact specification of infinite sets of graphs.
\acp{ReGaP} can be further extended with new wildcard types, such as optional nodes/edges and sequences/subgraphs with size limitations.

\section{\ac{ReGaP} Matching Encoding}
\label{sec:regap-enc}

Our approach reduces \ac{ReGaP} matching to an instance of \ac{SAT}.
First we detail the base encoding for the special case where $V_P$ does not contain wildcards.
The adaptations needed to support each wildcard type are explained in Sections~\ref{sec:one-plus-seq-enc} and~\ref{sec:one-plus-subgraph-enc}.
The full encoding relies on the implicit mapping that exists between the nodes on the left of a generalization rule and the respective generalized node on the right.
For example, in rule 1, the node $u$ is mapped to the any-1+ wildcard that is introduced in its place.
In rule 2, the node $u$ and the nodes mapped to $S$ become mapped to the new $S+$.

The base encoding is an adaptation of an encoding for maximum common subgraph available in the literature~\cite{malware-sig-synth,DBLP:conf/sigsoft/Terra-NevesNVRV21}.
It solves the original problem by mapping the nodes and edges of $G$ into those of $P$.
For simplicity, some constraints are shown as at-most-1 constraints, i.e. of the form $\sum_i l_i \le 1$, instead of clauses.
Note that these can be converted to \ac{CNF} by introducing the clause $(\neg l_i \vee \neg l_j)$ for each pair $i, j$ such that $i \neq j$, or by using one of many \ac{CNF} encodings available in the literature~\cite{DBLP:conf/sat/AnsoteguiM04,chen2010new,DBLP:journals/jar/FrischPDN05,klieber2007efficient,DBLP:conf/sat/Prestwich07}.

The following sets of Boolean variables are considered:
\begin{itemize}
    \item \textbf{Inclusion variables.}
    For each node $v_P \in V_P$, a variable $o_{v_P}$ is introduced to encode if some node of $V$ is mapped to $v_P$, i.e. if there exists a node $v \in V$ such that $f(v) = v_P$ (i.e. $o_{v_P} = 1$) or not (i.e. $o_{v_P} = 0$).
    \item \textbf{Mapping variables.}
    For each node pair $(v_P, v) \in V_P \times V$, a variable $m_{v_P, v}$ is used to encode if the node $v$ is mapped to $v_P$.
    If $f(v) = v_P$, then $m_{v_P, v} = 1$, otherwise $m_{v_P, v} = 0$.
    \item \textbf{Control-flow variables.}
    These variables are the analogous of the inclusion variables for edges.
    For each edge $(u_P, v_P) \in E_P$, a variable $c_{u_P, v_P}$ is used to encode if there exists an edge $(u, v) \in E$ mapped to $(u_P, v_P)$.
    If so, then $c_{u_P, v_P} = 1$, otherwise $c_{u_P, v_P} = 0$.
    $(u, v)$ is said to be mapped to $(u_P, v_P)$ if both $u$ and $v$ are mapped to $u_P$ and $v_P$ respectively.
\end{itemize}

The \ac{SAT} formula contains the following clauses:
\begin{itemize}
    \item \textbf{Inclusion clauses.}
    Ensure consistency between the inclusion and the mapping variables, i.e. for each node $v_P \in V_P$, if $o_{v_P} = 1$, then at least one of the $m_{v_P, v}$ must also be set to 1 for some $v \in V$, and vice-versa.
    \begin{equation}
        \bigwedge_{v_P \in V_P} \biggl( o_{v_P} \leftrightarrow \bigvee_{v \in V} m_{v_P, v} \biggr) \text{.}
    \end{equation}
    \item \textbf{One-to-one clauses.}
    Each node in $V$ must be mapped to at most one node in $V_P$ and vice-versa.
    %For each node $v_P \in V_P$, there exists at most one node $v \in V$ such that $f(v) = v_P$, and vice-versa:
    \begin{equation}
        \smashoperator[r]{\bigwedge_{v_P \in V_P}} \ \biggl( \smashoperator[r]{\sum_{v \in V}} m_{v_P, v} \le 1 \biggr) \wedge \smashoperator{\bigwedge_{v \in V}} \biggl( \smashoperator[r]{\sum_{v_P \in V_P}} m_{v_P, v} \le 1 \biggr) \text{.}
    \end{equation}
    \item \textbf{Control-flow consistency clauses.}
    Each edge in $E_P$ can only be mapped to edges that exist in $E$.
    More specifically, for each edge $(u_P, v_P) \in E_P$, if $(u, v) \notin E$, then either $u$ is not mapped to $u_P$ (i.e. $m_{u_P, u} = 0$), $v$ is not mapped to $v_P$ (i.e. $m_{v_P, v} = 0$), or no edge of $E$ is mapped to $(u_P, v_P)$ (i.e. $c_{u_P, v_P} = 0$).
    \begin{equation}\label{eq:ctrl-flow-consistency}
        \bigwedge_{(u_P, v_P) \in E_P} \smashoperator[r]{\bigwedge_{(u, v) \in (V \times V) \setminus E}} \left( \overline{m_{u_P, u}} \vee \overline{m_{v_P, v}} \vee \overline{c_{u_P, v_P}} \right) \text{.}
    \end{equation}
    \item \textbf{No spurious edge clauses.}
    If an edge $(u_P, v_P) \in E_P$ is mapped to some edge in $E$, then $u_P$ and $v_P$ must also be mapped to nodes of $V$.
    %An edge in $E$ can be mapped to some $(u_P, v_P) \in E_P$ only if there exists a node pair $(u, v) \in V \times V$ such that $f(u) = u_P$ and $f(v) = v_P$:
    \begin{equation}\label{eq:spurious}
        \smashoperator{\bigwedge_{(u_P, v_P) \in E_P}}  \left( c_{u_P, v_P} \rightarrow o_{u_P} \wedge  o_{v_P}  \right) \text{.}
    \end{equation}
    \item \textbf{Node isomorphism clauses.}
    All nodes in $V$ must be mapped to a node in $V_P$ and vice-versa.
    %For each node in $V_P$, we need to enforce that its inclusion variable is set to 1.
    %For each node in $V$, we must ensure that at least one of its mapping variables is set to 1.
    \begin{equation}\label{eq:node-iso}
        \smashoperator[r]{\bigwedge_{v_P \in V_P}} \left( o_{v_P} \right) \wedge \bigwedge_{v \in V} \biggl( \smashoperator[r]{\bigvee_{v_P \in V_P}} m_{v_P, v} \biggr) \text{.}
    \end{equation}
    \item \textbf{Edge isomorphism clauses.}
    All edges in $E$ must be mapped to an edge in $E_P$ and vice-versa.
    %For each edge in $E_P$, we only need to enforce that its control-flow variable is set to 1.
    For each edge $(u, v) \in E$, we must ensure that, given a pair of nodes $u_P, v_P$ of $V_P$ such that $(u_P, v_P) \notin E_P$, then either $u$ or $v$ is not mapped to $u_P$ or $v_P$ respectively.
    \begin{equation}\label{eq:edge-iso}
        \smashoperator[r]{\bigwedge_{(u_P, v_P) \in E_P}} \left( c_{u_P, v_P} \right) \wedge \smashoperator[l]{\bigwedge_{(u, v) \in E}} \smashoperator[r]{\bigwedge_{(u_P, v_P) \in (V_P \times V_P) \setminus E_P}} \left( \overline{m_{u_P, u}} \vee \overline{m_{v_P, v}} \right) \text{.}
    \end{equation}
\end{itemize}

\subsection{Sequence Wildcards}
\label{sec:one-plus-seq-enc}

In order to support any-0+-sequence wildcards, each $w \in W_P^{S*}$ must be expanded by replacing it with $k$ non-wildcard nodes $w^1, \dots, w^k$, as well as $k-1$ edges, one for each $(w^i, w^{i+1})$ such that $i \in \lbrace 1..k-1 \rbrace$.
The choice of value for $k$ is discussed at the end of this section.
We use $V_P^{\mathrm{exp}}(w) = \lbrace w^1, \dots, w^k \rbrace$ to denote the set of non-wildcard nodes added to replace $w$, and $E_P^{\mathrm{exp/mid}}(w) = \lbrace (w^1, w^2), \dots, (w^{k-1}, w^k) \rbrace$ to denote the set of edges added between the nodes of $V_P^{\mathrm{exp}}(w)$.
Each edge $(u_P, w) \in E_P$ is replaced by the edge $(u_P, w^1)$.
We use $E_P^{\mathrm{exp/in}}(w)$ to denote the set of edges added to replace each such $(u_P, w)$.
Similarly, each $(w, v_P) \in E_P$ is replaced by $k$ edges $(w^i, v_P)$, one for each $i \in \lbrace 1..k \rbrace$.
We use $E_P^{\mathrm{exp/out}}(w, v_P)$ to denote the set of all new edges added to replace $(w, v_P)$.
Additionally, we use $S_P(w) \subset V_P$ to denote the set of successors of $w$, i.e. $S_P(w) = \lbrace v_P \in V_P: (w, v_P) \in E_P \rbrace$, and $E_P^{\mathrm{exp/out}}(w)$ to denote the union of all sets $E_P^{\mathrm{exp/out}}(w, v_P)$, i.e. $E_P^{\mathrm{exp/out}}(w) = \bigcup_{v_P \in S_P(w)} E_P^{\mathrm{exp/out}}(w, v_P)$.

Extra edges $(u_P, v_P)$ are added from each predecessor $u_P$ of $w$ to each successor $v_P \in S_P(w)$.
We use $E_P^{\mathrm{exp/skip}}(u_P, w)$ to denote the set of edges added from $u_P$ to the nodes in $S_P(w)$.
Additionally, we use $B_P(w) \subset V_P$ to denote the set of predecessor nodes of $w$, i.e. $B_P(w) = \lbrace u_P \in V_P: (u_P, w) \in E_p \rbrace$, and $E_P^{\mathrm{exp/skip}}(w, v_P)$ to denote the set of edges added from the nodes in $B_P(w)$ to $v_P$.
We use $E_P^{\mathrm{exp/skip}}(w)$ to denote the union of all such sets, i.e. $E_P^{\mathrm{exp/skip}}(w) = \bigcup_{u_P \in B_P(w)} E_P^{\mathrm{exp/skip}}(u_P, w)$.
Lastly, we use $E_P^{\mathrm{exp}}(w)$ to denote all edges added when replacing $w$, i.e. $E_P^{\mathrm{exp}}(w) = E_P^{\mathrm{exp/in}}(w) \cup E_P^{\mathrm{exp/mid}}(w) \cup E_P^{\mathrm{exp/skip}}(w) \cup E_P^{\mathrm{exp/out}}(w)$.

Let $P^{\mathrm{exp}} = (V_P^{\mathrm{exp}}, E_P^{\mathrm{exp}})$ denote the graph that results from the expansion.
The encoding is built using $P^{\mathrm{exp}}$ instead of $P$, with the following changes:
\begin{itemize}
    \item \textbf{Node isomorphism clauses.}
    For each wildcard $w \in W_P^{S*}$, the nodes in $V_P^{\mathrm{exp}}(w)$ are optional and thus must be excluded from equation~\eqref{eq:node-iso} as follows:
    \begin{equation}
        \label{eq:seq-one-plus-node-iso}
        \smashoperator[r]{\bigwedge_{v_P \in V_P^{\mathrm{exp}} \setminus \bigcup_{w \in V_P^{S*}} V_P^{\mathrm{exp}}(w)}} \left( o_{v_P} \right) \quad \; \wedge \smashoperator{\bigwedge_{v \in V}} \biggl( \smashoperator[r]{\bigvee_{v_P \in V_P^{\mathrm{exp}}}} m_{v_P, v} \biggr) \text{.}
    \end{equation}
    \item \textbf{Edge isomorphism clauses.}
    Analogously, the edges in $E_P^{\mathrm{exp}}(w)$ must be excluded from equation~\eqref{eq:edge-iso} as follows:
    \begin{multline}\label{eq:seq-zero-plus-edge-iso}
        \smashoperator[r]{\bigwedge_{(u_P, v_P) \in E_P^{\mathrm{exp}} \setminus \bigcup_{w \in W_P^{S*}} E_P^{\mathrm{exp}}(w)}} \left( c_{u_P, v_P} \right) \wedge \\ \smashoperator[l]{\bigwedge_{(u, v) \in E}} \smashoperator[r]{\bigwedge_{(u_P, v_P) \in (V_P^{\mathrm{exp}} \times V_P^{\mathrm{exp}}) \setminus E_P^{\mathrm{exp}} }} \left( \overline{m_{u_P, u}} \vee \overline{m_{v_P, v}} \right) \text{.}
    \end{multline}
    However, extra clauses are necessary to ensure that an optional edge $(u_P, v_P)$ is mapped to the edge $(u, v) \in E$ when $u$ and $v$ are mapped to $u_P$ and $v_P$ respectively.
    \begin{equation}\label{eq:seq-one-plus-ctrl-flow-ensurance}
        \bigwedge_{(u_P, v_P) \in E_P^{\mathrm{exp}}} \smashoperator[r]{\bigwedge_{(u, v) \in E}} \left( m_{u_P, u} \wedge m_{v_P, v} \rightarrow c_{u_P, v_P} \right) \text{.}
    \end{equation}
    \item \textbf{Sequence clauses.}
    A sequence node $w^i$ can be mapped to some node in $V$ only if each of the sequence nodes that precede $w^i$ have some node of $V$ mapped to them.
    %It is sufficient to enforce that $o_{w^i} = 1$ implies that $o_{w^{i-1}} = 1$.
    \begin{equation}\label{eq:seq-one-plus-seq}
        \bigwedge_{w \in W_P^{S*}} \smashoperator[r]{\bigwedge_{w^i \in V_P^{\mathrm{exp}}(w), i \ge 2}} \left( o_{w^i} \rightarrow o_{w^{i-1}} \right) \text{.}
    \end{equation}
    \item \textbf{Incoming any-0+-sequence control-flow clauses.}
    If a node in $V$ is mapped to a wildcard $w \in W_P^{S*}$, then each incoming edge of $w$ must be mapped to an edge in $E$.
    \begin{equation}\label{eq:seq-zero-plus-in-ctrl-flow}
        \bigwedge_{w \in W_P^{S*}} \smashoperator[r]{\bigwedge_{(u_P, v_P) \in E_P^{\mathrm{exp/in}}(w)}} \left( o_{w^1} \rightarrow c_{u_P, v_P} \right) \text{.}
    \end{equation}
    \item \textbf{Outgoing any-0+ control-flow clauses.}
    Analogous of equation~\eqref{eq:seq-zero-plus-in-ctrl-flow} but for the outgoing edges of $w$.
    \begin{equation}\label{eq:seq-one-plus-outgoing-atleast1}
        \bigwedge_{w \in W_P^{S*}} \bigwedge_{v_P \in S_P(w)} \\ \biggl( o_{w^1} \rightarrow \quad \smashoperator{\bigvee_{(u_P, v_P) \in E_P^{\mathrm{exp/out}}(w, v_P)}} c_{u_P, v_P} \quad \biggr) \text{.}
    \end{equation}
    Note that, while $w^1$ must always be the first sequence node mapped to some node of $V$, the last such sequence node $w^l$ can vary depending on the number $l$ of nodes of $V$ mapped to $w$.
    Only the outgoing edges of $w^l$ can be mapped to some edge in $E$.
    \begin{equation}\label{eq:seq-one-plus-outgoing-last}
        \bigwedge_{w \in W_P^{S*}} \smashoperator[r]{\bigwedge_{(w^i, v_P) \in E_P^{\mathrm{exp/out}}(w), i \le k-1}} \left( o_{w^{i+1}} \rightarrow \overline{c_{w^i, v_P}} \right) \text{.}
    \end{equation}
    \item \textbf{Skip any-0+-sequence control-flow clauses.}
    For each wildcard $w \in W_P^{S*}$, if no node in $V$ is mapped to $w$ and $w$ has at least one successor, then each predecessor $u_P$ of $w$ must have at least one of the edges that connect $u_P$ to one of the successors of $w$ mapped to some edge in $E$.
    \begin{equation}\label{eq:seq-zero-plus-outgoing-atleast1-skip}
        \bigwedge_{w \in W_P^{S*}, \left| S_P(w) \right| > 0} \bigwedge_{u_P \in B_P(w)} \biggl( \overline{o_{w^1}} \rightarrow \quad \smashoperator{\bigvee_{(u_P, v_P) \in E_P^{\mathrm{exp/skip}}(u_P, w)}} c_{u_P, v_P} \quad \biggr) \text{.}
    \end{equation}
    Analogous of equation~\eqref{eq:seq-zero-plus-outgoing-atleast1-skip} but for the successors.
    \begin{equation}\label{eq:seq-zero-plus-outgoing-atleast1-skip-sucs}
        \bigwedge_{w \in W_P^{S*}, \left| B_P(w) \right| > 0} \bigwedge_{v_P \in S_P(w)} \biggl( \overline{o_{w^1}} \rightarrow \quad \smashoperator{\bigvee_{(u_P, v_P) \in E_P^{\mathrm{exp/skip}}(w, v_P)}} c_{u_P, v_P} \quad \biggr) \text{.}
    \end{equation}
    Lastly, if a node in $V$ is mapped to $w$, then the edges in $E$ cannot be mapped to an edge in $E_P^{\mathrm{exp/skip}}(w)$.
    \begin{equation}
        \bigwedge_{w \in W_P^{S*}} \smashoperator[r]{\bigwedge_{(u_P, v_P) \in E_P^{\mathrm{exp/skip}}(w)}} \left( o_{w^1} \rightarrow \overline{c_{u_P, v_P}} \right) \text{.}
    \end{equation}
\end{itemize}

Note that the choice of $k$ must ensure that $P^{\mathrm{exp}}$, together with the aforementioned changes to the encoding, retains the same semantics as $P$.
One (naive) solution is to set $k = \left| V \right|$.
Moreover, for the sake of simplicity, the encoding, as described, assumes that $P$ does not contain edges between wildcards.
The expansion of such wildcards is actually an iterative process.
Therefore, an edge $(w, w') \in E_P$ between a pair of wildcards $(w, w') \in W_P^{S*} \times W_P^{S*}$ ends up being replaced by $k$ $(w^i, w'^1)$ edges from each node $w^i \in V_P^{\mathrm{exp}}(w)$ to the first non-wildcard node $w'^1$ introduced to replace $w'$.
Additional edges must also be added from $w^1, \dots, w^k$ and the predecessors of $w$ to the successors of $w'$.

Given the above encoding, any-1+-sequence wildcards are supported by replacing each such $w \in W_P^{S+}$ with a non-wildcard node $w^1$, an any-0+-sequence $w'$ and the edge $(w^1, w')$, and by setting the destination (source) of all incoming (outgoing) edges of $w$ to $w^1$ ($w'$).

\subsection{Any-1+-Subgraph Wildcards}
\label{sec:one-plus-subgraph-enc}

In order to support any-1+-subgraph wildcards, each $w \in W_P^{G+}$ is replaced by a copy of $G$, i.e. $w$ is replaced by a $w^v$ node for each $v \in V$ and a $(w^u, w^v)$ edge for each $(u, v) \in E$.
As in Section~\ref{sec:one-plus-seq-enc}, $V_P^{\mathrm{exp}}(w)$ ($E_P^{\mathrm{exp/mid}}(w)$) denotes the set of node (edge) copies created to replace $w$.
Each edge $(u_P, w) \in E_p$ is replaced by $\left| V \right|$ edges $(u_P, w^v)$ from $u_P$ to each $w^v \in V_P^{\mathrm{exp}}(w)$.
We use $E_P^{\mathrm{exp/in}}(u_P, w)$ to denote the set of new edges that replace $(u_P, w)$.
Similarly, each edge $(w, v_P) \in E_P$ is replaced by $\left| V \right|$ edges $(w^v, v_P)$ from each $w^v \in V_P^{\mathrm{exp}}(w)$ to $v_P$.
$E_P^{\mathrm{exp/out}}(w, v_P)$ denotes the set of new edges that replace $(w, v_P)$.

The encoding in Section~\ref{sec:one-plus-seq-enc} can be adapted to support any-1+-subgraph wildcards by replacing $W_P^{S*}$ with $W_P^{S*} \cup W_P^{G+}$ in equation~\eqref{eq:seq-one-plus-node-iso}, equation~\eqref{eq:seq-zero-plus-edge-iso} and equation~\eqref{eq:seq-one-plus-ctrl-flow-ensurance}, plus the following new clauses:
\begin{itemize}
    \item \textbf{Any-1+-subgraph inclusion clauses.}
    At least one node in $V$ must be mapped to each wildcard $w \in W_P^{G+}$.
    \begin{equation}\label{eq:subgraph-one-plus-inclusion}
        \bigwedge_{w \in W_P^{G+}} \biggl( \smashoperator[r]{\bigvee_{v_P \in V_P^{\mathrm{exp}}(w)}} o_{v_P} \biggr) \text{.}
    \end{equation}
    \item \textbf{Incoming any-1+-subgraph control-flow clauses.}
    Each incoming edge of $w \in W_P^{G+}$ must be mapped to some edge in $E$.
    \begin{equation}\label{eq:subgraph-one-plus-in-ctrl-flow}
        \bigwedge_{w \in W_P^{G+}} \bigwedge_{u_P \in B_P(w)} \biggl( \smashoperator[r]{\bigvee_{(u_P, v_P) \in E_P^{\mathrm{exp/in}}(u_P, w)}} c_{u_P, v_P} \ \ \; \biggr) \text{.}
    \end{equation}
    \item \textbf{Outgoing any-1+ control-flow clauses.}
    Analogous of equation~\eqref{eq:subgraph-one-plus-in-ctrl-flow} for the outgoing edges.
    \begin{equation}\label{eq:subgraph-one-plus-outgoing-atleast1}
        \bigwedge_{w \in W_P^{G+}} \bigwedge_{v_P \in S_P(w)} \biggl( \smashoperator[r]{\bigvee_{(u_P, v_P) \in E_P^{\mathrm{exp/out}}(w, v_P)}} c_{u_P, v_P} \ \ \; \, \biggr) \text{.}
    \end{equation}
\end{itemize}

\subsection{Any-0+-subgraph Wildcards}
\label{sec:zero-plus-subgraph-enc}

To support any-0+-subgraph wildcards, each such $w \in W_P^{G*}$ is replaced by new nodes and edges as described in Section~\ref{sec:one-plus-subgraph-enc}, plus an extra set of edges from each predecessor $u_P \in B_P(w)$ to each successor $v_P \in S_P(w)$ as described in Section~\ref{sec:one-plus-seq-enc}.

The encoding in section~\ref{sec:one-plus-subgraph-enc} can be adapted to support any-0+-subgraph wildcards by replacing $W_P^{S*} \cup W_P^{G+}$ with $W_P$ in equation~\eqref{eq:seq-one-plus-node-iso}, equation~\eqref{eq:seq-zero-plus-edge-iso} and equation~\eqref{eq:seq-one-plus-ctrl-flow-ensurance}, and by adding the following clauses:
\begin{itemize}
    \item \textbf{Incoming any-0+-subgraph control-flow clauses.}
    These clauses are similar to those of equation~\eqref{eq:subgraph-one-plus-in-ctrl-flow}, with the exception that the constraint also allows the edges in $E_P^{\mathrm{exp/skip}}(u_P, w)$, in addition to those in $E_P^{\text{E\text{in}}}(u_P, w)$, and only applies if at least one node in $V$ is mapped to $w$.
    \begin{equation}\label{eq:subgraph-zero-plus-in-ctrl-flow}
        \bigwedge_{w \in V_P^{G*}} \bigwedge_{u_P \in B_P(w)} \biggl( \smashoperator[r]{\bigvee_{v_P \in V_P^{\mathrm{exp}}(w)}} o_{v_P} \quad  \rightarrow \quad \smashoperator{\bigvee_{\substack{(u_P', v_P') \in\\ E_P^{\mathrm{exp/in}}(u_P, w) \cup E_P^{\mathrm{exp/skip}}(u_P, w)}}} c_{u_P', v_P'} \biggr) \text{.}
    \end{equation}
    However, additional clauses are necessary to ensure that at least one of the incoming edges of $w$ is mapped to some edge in $E$.
    \begin{equation}\label{eq:subgraph-zero-plus-in-ctrl-flow-at-least1}
        \bigwedge_{w \in V_P^{G*}} \biggl( \smashoperator[r]{\bigvee_{v_P \in V_P^{\mathrm{exp}}(w)}} o_{v_P} \quad \rightarrow \quad  \smashoperator{\bigvee_{(u_P', v_P') \in E_P^{\mathrm{exp/in}}(w)}} c_{u_P', v_P'} \, \biggr) \text{.}
    \end{equation}
    \item \textbf{Outgoing any-0+-subgraph control-flow clauses.}
    Analogous of equation~\eqref{eq:subgraph-zero-plus-in-ctrl-flow} and equation~\eqref{eq:subgraph-zero-plus-in-ctrl-flow-at-least1} for the outgoing edges.
    These clauses share the same aforementioned similarities with those of equation~\eqref{eq:subgraph-one-plus-outgoing-atleast1}.
    \begin{equation}\label{eq:seq-zero-plus-subgraph-outgoing-atleast1-out}
        \bigwedge_{w \in V_P^{G*}} \bigwedge_{v_P \in S_P(w)} \biggl( \smashoperator[r]{\bigvee_{u_P \in V_P^{\mathrm{exp}}(w)}} o_{u_P} \quad  \rightarrow \quad \smashoperator{\bigvee_{\substack{(u_P', v_P') \in\\ E_P^{\mathrm{exp/out}}(w, v_P) \cup E_P^{\mathrm{exp/skip}}(w, v_P)}}} c_{u_P', v_P'} \biggr) \text{.}
    \end{equation}
    \begin{equation}
        \bigwedge_{w \in V_P^{G*}} \biggl( \smashoperator[r]{\bigvee_{u_P \in V_P^{\mathrm{exp}}(w)}} o_{u_P} \quad \rightarrow \quad  \smashoperator{\bigvee_{(u_P', v_P') \in E_P^{\mathrm{exp/out}}(w)}} c_{u_P', v_P'} \: \biggr) \text{.}
    \end{equation}
    \item \textbf{Skip any-0+-subgraph control-flow clauses.}
    Analogous of equation~\eqref{eq:seq-zero-plus-outgoing-atleast1-skip} and equation~\eqref{eq:seq-zero-plus-outgoing-atleast1-skip-sucs} for any-0+-subgraph wildcards.
    \begin{equation}\label{eq:subgraph-zero-plus-outgoing-atleast1-skip}
        \bigwedge_{\substack{w \in V_P^{G*} \\ \left| S_P(w) \right| > 0}} \bigwedge_{u_P \in B_P(w)} \biggl( \smashoperator[r]{\bigwedge_{v_P \in V_P^{\mathrm{exp}}(w)}} \overline{o_{v_P}} \quad \, \rightarrow \quad \, \smashoperator{\bigvee_{(u_P', v_P') \in E_P^{Eskip}(u_P, w)}} c_{u_P', v_P'} \ \ \, \biggr) \text{.}
    \end{equation}
    \begin{equation}\label{eq:subgraph-zero-plus-outgoing-atleast1-skip-suc}
        \bigwedge_{\substack{w \in V_P^{G*} \\ \left| B_P(w) \right| > 0}} \bigwedge_{v_P \in S_P(w)} \biggl( \smashoperator[r]{\bigwedge_{u_P \in V_p^{\mathrm{exp}}(w)}} \overline{o_{u_P}} \quad \, \rightarrow \quad \, \smashoperator{\bigvee_{(u_P', v_P') \in E_P^{Eskip}(w, v_P)}} c_{u_P', v_P'} \ \ \, \biggr) \text{.}
    \end{equation}
\end{itemize}

\section{Attributed \ac{ReGaP} Matching}
\label{sec:attr-regap}

In the attributed \ac{ReGaP} matching problem, $G$ is an attributed graph and $P$ defines constraints over the attributes of the nodes/edges of $G$.
There are 3 types of constraints:
\begin{itemize}
    \item \textbf{Node constraints.} Each node $v_P \in V_P$ is assigned a node constraint $\phi_{v_P}$ over the attributes $A_V$ of the nodes in $V$.
    Given a node $v \in V$, we use $\phi_{v_P}(v) = 1$ ($\phi_{v_p}(v) = 0$) to denote that $v$ satisfies (does not satisfy) $\phi_{v_P}$.
    \item \textbf{Edge constraints.} Analogously, an edge constraint $\phi_{(u_P, v_P)}$ is associated with each edge $(u_P, v_P) \in E_P$ and $\phi_{(u_P, v_P)}((u, v))$ denotes if the edge $(u, v) \in E$ satisfies $\phi_{(u_P, v_P)}$.
    \item \textbf{Node pair relation constraints.} A node pair relation constraint $\psi_{u_P, v_P}$ is associated with each node pair $u_P, v_P \in V_p$ such that $u_P \neq v_P$.
    Note that the existence of a node pair relation constraint between $u_P$ and $v_P$ does not imply that the edge $(u_P, v_P)$ exists.
\end{itemize}

The definition for the attributed problem must ensure that these constraints are satisfied.
We assume that node and node pair relation constraints cannot be associated with wildcards.

\begin{definition}\label{def:attr-regap-match}
Let $P = (V_P, E_P)$ be a \ac{ReGaP} and $G$ an attributed graph.
$P$ is said to match $G$ if and only if there exists a sequence of generalization rules transforming $G$ into $G' = (V', E')$ such that there exists a bijective mapping $f: V_P \leftrightarrow V'$ that satisfies the following conditions:
\begin{enumerate}
    \item for all $(u, v) \in E_P$, $(f(u), f(v)) \in E'$ and vice-versa.
    \item for all $w \in W_P$, $f(w)$ is a wildcard of the same type.
    \item\label{enum:node-constr-sat} for all $v \in V_P \setminus W_P$, $\phi_v(f(v)) = 1$.
    \item for all $(u, v) \in E_P$, $\phi_{(u, v)}((f(u), f(v))) = 1$.
    \item for all $u, v \in (V_P \setminus W_P) \times (V_P \setminus W_P)$ such that $u \neq v$, $\psi_{u, v}(f(u), f(v)) = 1$.
\end{enumerate}
\end{definition}

\subsection{Encoding}

This section describes how to adapt the encoding in Section~\ref{sec:regap-enc} for attributed matching.
First, one must set the constraints for the extra nodes/edges added by wildcard expansion.
Given a wildcard $w$ of any type, the node constraint for each $w^i \in V_P^{\mathrm{exp}}(w)$ is set to $\phi_{w^i}(v) \equiv 1$, i.e. any node of $V$ can be mapped to $w^i$.
The same applies to the edge constraints for each edge in $E_P^{\mathrm{exp/mid}}(w)$.
Given a predecessor $u_P \in B_P(w)$, the edge constraint for each $(u_P, w^i) \in E_P^{\mathrm{exp/in}}(u_P, w)$ is set to $\phi_{(u_P, w^i)}((u, v)) \equiv \phi_{(u_P, w)}((u, v))$.
If $w$ is an any-0+ wildcard, the constraints for the edges in $E_P^{\mathrm{exp/skip}}(u_P, w)$ are set in the same way.
Lastly, given a successor $v_P \in S_P(w)$, the edge constraint for each $(w^i, v_P) \in E_P^{\mathrm{exp/out}}(w, v_P)$ is $\phi_{(w^i, v_P)}((u, v)) \equiv \phi_{(w, v_P)}((u, v))$.

The following additional clauses are necessary:
\begin{itemize}
    \item \textbf{Node constraint consistency clauses.}
    The node constraints in $P$ must be satisfied.
    More specifically, if a node $v \in V$ does not satisfy the node constraint of a node $v_P \in V_P$, then $v$ cannot be mapped to $v_P$.
    \begin{equation}\label{eq:node-constr-consistency}
        \bigwedge_{v_P \in V_P} \smashoperator[r]{\bigwedge_{v \in V, \phi_{v_P}(v) = 0}} \left( \overline{m_{v_P, v}} \right) \text{.}
    \end{equation}
    \item \textbf{Node pair relation constraint consistency clauses.}
    The node pair relation constraints in $P$ must be satisfied.
    More specifically, for each pair of nodes $u_P, v_P$ of $V_P$ and $u, v$ of $V$, if $u$ is mapped to $u_P$ and the pair $u, v$ does not satisfy the node pair relation constraint of $u_P, v_P$, then $v$ cannot be mapped to $v_P$.
    \begin{equation}
        \bigwedge_{\substack{(u_P, v_P) \in V_P \times V_P \\ u_P \neq v_P}}\\ \bigwedge_{u \in V} \biggl( m_{u_P, u} \rightarrow \smashoperator{\bigvee_{\substack{v \in V \\ u \neq v \wedge \psi_{u_P, v_P}(u, v) = 1}}} m_{v_P, v} \biggr) \text{.}
    \end{equation}
\end{itemize}
Edge constraint satisfaction is ensured by changing the subscript of the inner conjunctions in equation~\eqref{eq:ctrl-flow-consistency} and equation~\eqref{eq:seq-zero-plus-edge-iso} to exclude edges such that $\phi_{(u_P, v_P)}((u, v)) = 0$.

\subsection{Node Merging}

Wildcard expansion (see Sections~\ref{sec:one-plus-seq-enc} and~\ref{sec:one-plus-subgraph-enc}) can have a severe impact in the size of the encoding, and thus the performance of the \ac{SAT} solver.
To mitigate this, we propose a sound and complete procedure that merges sequences of nodes in $G$ that do not satisfy any node constraints in $P$, since such nodes can only be mapped to wildcards, regardless of the wildcard type, as long as $P$ does not contain edges between wildcards.

\begin{proposition}\label{prop:node_merging}
    Consider an attributed graph $G = (V, E)$ and a ReGaP $P = (V_P, E_P)$ with no edges between wildcards, i.e. $\left| W_P \cap \lbrace u_P, v_P \rbrace \right| \le 1$ for all $(u_P, v_P) \in E_P$, and an edge $(u, v) \in E$ such that: for all $v_P \in V_P \setminus W_P$, $\phi_{v_P}(u) = \phi_{v_P}(v) = 0$; for all $(u', v) \in E$, $u' = u$; and, for all $(u, v') \in E$, $v' = v$.
    Let $G' = (V', E')$ be an attributed graph such that: 
    \begin{itemize}
    \item $V' = V \setminus \lbrace u \rbrace$; $A_V'(v') = A_V(v')$ for all $v' \in V'$;
    \item  $E' = \left[ E \setminus \left( \lbrace (u, v) \rbrace \cup \lbrace (u', u): (u', u) \in E \rbrace \right) \right] \cup \lbrace (u', v): (u', u) \in E \rbrace$;
    \item  $A_E'((u', v')) = A_E((u', v'))$ for all $(u', v') \in E \cap E'$;
    \item $A_E'((u', v)) = A_E((u', u))$ for all $(u', u) \in E$.
    \end{itemize}
    $P$ matches $G$ if and only if $P$ matches $G'$. 
\end{proposition}

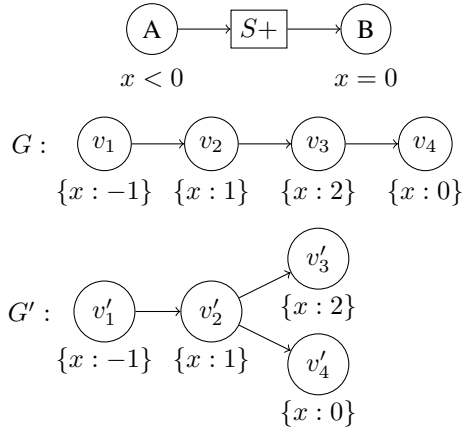
\begin{figure}[t]
    \centering
    \begin{tikzpicture}
      \node[circle, draw] (v1) {$v_1$};
      \node[below of=v1, yshift=1em] {$\lbrace x: -1 \rbrace$};
      \node[circle, right of=v1, draw, xshift=1.2em] (v2) {$v_2$};
      \node[below of=v2, yshift=1em] {$\lbrace x: 1 \rbrace$};
      \node[circle, right of=v2, draw, xshift=1.2em] (v3) {$v_3$};
      \node[below of=v3, yshift=1em] {$\lbrace x: 2 \rbrace$};
      \node[circle, right of=v3, draw, xshift=1.2em] (v4)  {$v_4$};
      \node[below of=v4, yshift=1em] {$\lbrace x: 0 \rbrace$};
      
      \draw[->] (v1) -- (v2);
      \draw[->] (v2) -- (v3);
      \draw[->] (v3) -- (v4);
    
      \node[circle, above of=v1, draw, yshift=1.5em, xshift=1.8em] (A) {A};
      \node[below of=A, yshift=1em] {$x < 0$};
      \node[right of=A, draw, xshift=1.2em] (S) {$S+$};
      \node[circle, right of=S, draw, xshift=1.2em] (B) {B};
      \node[below of=B, yshift=1em] {$x = 0$};
      \node[left of=v1, xshift=0em] {$G:$};
      
      \draw[->] (A) -- (S);
      \draw[->] (S) -- (B);
      
      \node[circle, below of=v1, yshift=-3.5em, draw] (v1p) {$v_1'$};
      \node[below of=v1p, yshift=1em] {$\lbrace x: -1 \rbrace$};
      \node[circle, right of=v1p, draw, xshift=1.2em] (v2p) {$v_2'$};
      \node[below of=v2p, yshift=1em] {$\lbrace x: 1 \rbrace$};
      \node[circle, right of=v2p, draw, xshift=1.2em, yshift=2em] (v3p) {$v_3'$};
      \node[below of=v3p, yshift=1em] {$\lbrace x: 2 \rbrace$};
      \node[circle, right of=v2p, draw, xshift=1.2em, yshift=-2em] (v4p)  {$v_4'$};
      \node[below of=v4p, yshift=1em] {$\lbrace x: 0 \rbrace$};
      \node[left of=v1p, xshift=0em] {$G':$};
      
      \draw[->] (v1p) -- (v2p);
      \draw[->] (v2p) -- (v3p);
      \draw[->] (v2p) -- (v4p);
    \end{tikzpicture}
    \caption{An example of an attributed \ac{ReGaP} (top), an attributed graph $G$ that matches the \ac{ReGaP} and an attributed graph $G'$ that does not match.}
    \label{fig:merge_example}
\end{figure}

\begin{proof}
    Consider that $P$ matches $G$.
    Then, there exists a sequence of generalization rules $r_1 \dots r_n$ transforming $G$ into $G^R$ that satisfies the conditions of Definition~\ref{def:attr-regap-match}.
    $\phi_{v_P}(u) = 0$ for all $v_P \in V_P$ implies that there must exist $r_i$ in $r_1 \dots r_n$ that discards $u$, otherwise condition~\ref{enum:node-constr-sat} of Definition~\ref{def:attr-regap-match} would not be satisfied.
    The following scenarios are possible:
    \begin{enumerate*}[label=(\arabic*)]
        \item\label{enum:u-to-any-1p} $r_i$ transforms $u$ into an any-1+ using rule 1;
        \item\label{enum:u-merge-any-seq} $r_i$ merges $u$ with an any-1+-sequence using rule 2.
    \end{enumerate*}

    Lets consider scenario~\ref{enum:u-to-any-1p}.
    $\phi_{v_P}(v) = 0$ for all $v_P \in V_P$ implies that there must exist $r_j$ in $r_1 \dots r_n$, $j \neq i$, that discards $v$.
    As with $r_i$, $r_j$ must:
    \begin{enumerate*}[label=\alph*)]
        \item\label{enum:v-rule1} transform $v$ into an any-1+ using rule 1, or
        \item\label{enum:v-rule2} merge $v$ with an any-0+-sequence using rule 2.
    \end{enumerate*}
    Note that $\left| W_P \cap \lbrace u_P, v_P \rbrace \right| \le 1$ for all $(u_P, v_P) \in E_P$, i.e. no edge exists between two wildcards in $P$.
    Therefore, there must exist $r_k$ in $r_1 \dots r_n$, $k > j$, that uses rule 6 to merge the any-1+ generated from $u$ with the any-1+ generated from $v$.
    Assuming scenario~\ref{enum:v-rule1} and that $r$ transforms $v$ into an any-1+-subgraph, we have that $r_1 \dots r_{i-1} r_{i+1} \dots r_{j-1} r r_{j+1} \dots r_{k-1} r_{k+1} \dots r_n$ transforms $G'$ into $G^R$.
    Note that we assume $j > i$, but the same also applies for $j < i$.
    The same applies to scenario~\ref{enum:v-rule2}, the difference being that $r_j$ is replaced with an application of rule 1 to transform $v$ into an any-1+-subgraph, an application of rule 3 to add an any-0+ between the transformed $v$ and its successor and two applications of rule 6 to merge the transformed $v$ with the any-0+ and its successor.
    In scenario~\ref{enum:u-merge-any-seq}, we have that $r_1 \dots r_{i-1} r_{i+1} \dots r_k$ transforms $G'$ into $G^R$.
    Therefore, there exists a sequence of generalization rules transforming $G'$ into $G_R$, thus $P$ matches $G'$.

    Now we prove that if $P$ matches $G'$, then $P$ matches $G$.
    It suffices to prove that all generalizations of $G'$ that discard $v$ are also generalizations of $G$.
    Let $r_1 \dots r_k$ be a sequence of generalization rules that transforms $G'$ into $G^R = (V^R, E^R)$ such that $v \notin V^R$.
    Then, there exists $r_i$ in $r_1 \dots r_k$ that discards $v$.
    The following scenarios are possible:
    \begin{enumerate*}[label=(\arabic*)]
        \item\label{enum:v-to-any-1p-seq} $r_i$ transforms $v$ into an any-1+-sequence using rule 1;
        \item\label{enum:v-to-any-1p-subgraph} $r_i$ transforms $v$ into an any-1+-subgraph using rule 1;
        \item\label{enum:v-merge-any-seq} $r_i$ merges $v$ with an any-1+-sequence using rule 2.
    \end{enumerate*}

    Lets consider scenario~\ref{enum:v-to-any-1p-seq}.
    Let $r$ be an application of rule 2 on $G$ that merges $u$ with a transformed $v$ and $r_j$, $j > i$, the first application of rule 2 in $r_1 \dots r_n$.
    We have that $r_1 \dots r_{j-1} r r_j \dots r_n$ transforms $G$ into $G^R$.
    The same applies to scenario~\ref{enum:v-to-any-1p-subgraph}, the difference being that $r$ must use rule 1 instead to transform $u$ into an any-1+ and that an additional application of rule 6 is required to merge the transformed $u$ and $v$.
    For scenario~\ref{enum:v-merge-any-seq}, we have that $r_1 \dots r_i r r_{i+1} \dots r_n$ transforms $G$ into $G^R$.
    Therefore, all generalizations of $G'$ that discard $v$ are also generalizations of $G$, and thus $P$ matches $G$.
\end{proof}

\begin{example}
    Consider the example \ac{ReGaP} and attributed graphs in Figure~\ref{fig:merge_example}.
    Node $v_1$ satisfies the node constraint of $A$, while $v_4$ satisfies the constraint of $B$.
    However, $v_2$ and $v_3$ do not satisfy either constraint.
    Therefore, we can safely merge $v_2$ and $v_3$ into a single node.
    On the other hand, $v_1'$ and $v_4'$ also satisfy the node constraints for $A$ and $B$ respectively, while $v_2'$ and $v_3'$ do not.
    However, $v_2'$ and $v_3'$ cannot be merged because there exists an edge from $v_2'$ to $v_4'$, thus $v_2'$ and $v_3'$ cannot be mapped to the same sequence wildcard.
\end{example}

Based on Proposition~\ref{prop:node_merging}, if $P$ does not contain edges between wildcard nodes, we apply a preprocessing step that repeatedly transforms $G$ into $G'$ until no more edges $(u, v)$ exist in $E$ satisfying the respective criteria.

\section{Experimental Evaluation}
\label{sec:eval}

This section evaluates the performance of the SAT-based approach for \ac{ReGaP} matching.
For the attributed graphs, we used a collection of control-flow graphs extracted from the Python code snippets in the CodeSearchNet dataset~\cite{DBLP:journals/corr/abs-1909-09436}.
The \acp{ReGaP} used in the evaluation replicate the kind of bad code patterns that are integrated in the AI Mentor Studio~\cite{aimentorstudio} code analysis engine for the OutSystems visual programming language, which uses \acp{ReGaP} as a formalism for specifying such patterns.
One concrete example of a bad performance pattern that occurs frequently in OutSystems is a database query Q1, followed by a loop that iterates the output of Q1 and performs another query Q2 with a filter by the current record of Q1.
Typically, Q2 can be merged with Q1 through a join condition, resulting in just $1$ query instead of $N+1$, where $N$ is the number of records returned by Q1.

\begin{table}[t]
\caption{Statistics on the benchmark set and \acp{ReGaP}. The G*, G+ and S* lines correspond to the number of any-0+-subgraph, any-1+-subgraph and any-0+-sequence wildcards respectively.}
\label{tb:inst-stats}
\centering
\begin{tabular}{l|c|c|c|c|}
\cline{2-5}
                                & min  & max  & median & mean  \\ \hline
\multicolumn{1}{|l|}{\#\acp{ReGaP}} & \multicolumn{4}{c|}{13}                                                                    \\ \hline
\multicolumn{1}{|l|}{\#G*}  & 0	& 5 &	3  &	2.6 \\ \hline
\multicolumn{1}{|l|}{\#G+}  & 0	& 3 &	0 &	0.3 \\ \hline
\multicolumn{1}{|l|}{\#S*}  & 0	& 1 &	0 &	0.1 \\ \hline
\multicolumn{1}{|l|}{\#Nodes Constr}  & 0	& 5 &	3  &	2.8 \\ \hline
\multicolumn{1}{|l|}{\#Edge Constr}  & 1	& 8 &	4  &	3.9 \\ \hline
\multicolumn{1}{|l|}{\#NodePair Rel. Constr}  & 0	& 1 &	0 &	0.3 \\ \hline \hline
\multicolumn{1}{|l|}{\#Instances} & \multicolumn{4}{c|}{946 556}                                                                    \\ \hline

\multicolumn{1}{|l|}{\#Nodes per instance}  & 15 & 1 070 &	21	& 26.6  \\ \hline
\multicolumn{1}{|l|}{\#Edges per instance}  & 14 & 1 335 &	24 & 30.5 \\ \hline
\end{tabular}
\end{table}

The graph dataset and \acp{ReGaP} are publicly available\footnote{https://github.com/MiguelTerraNeves/regap}, plus an executable that can be used to replicate the results presented in this evaluation.
13 \acp{ReGaP} are considered in this evaluation, containing up to 5 wildcards.
We consider only the 72 812 graphs with at least 15 nodes because the \ac{ReGaP} matcher was always able to solve the instances with smaller ones in under 7 seconds, resulting in a total of 946 556 instances.
The maximum number of nodes is 1070, with a median of 21.
Out of the 946 556 instances, 130 136 are know to be satisfiable, 791 426 are unsatisfiable, and 24 994 are unknown.
Table~\ref{tb:inst-stats} summarizes several statistics regarding the number of graphs and nodes/edges in this dataset.

\ac{SAT} formulas were solved using PySAT (version 0.1.7.dev21)~\cite{pysat}, configured to use the Glucose solver (version 4.1)~\cite{glucose} with default settings.
All experiments were run once with a timeout of 60 seconds\footnote{The 60 seconds timeout is what is considered acceptable in the context of AI Mentor Studio.} on an AWS m5a.24xlarge instance with 384 GB of RAM.
Different experiments were split across 84 workers running in parallel, each running Glucose sequentially.

We aim to answer the following research questions regarding the performance of the proposed approach:
\begin{enumerate*}
    \item[\textbf{R1:}] What is the impact of the node merging step?
    \item[\textbf{R2:}] What is the impact of the graph size?
    \item[\textbf{R3:}] What is the impact of the number of wildcards?
\end{enumerate*}

Recall from Section~\ref{sec:intro} that, although \acp{ReGaP} and regular-path queries are related, there exist structures expressible with \acp{ReGaP} that are not expressible using regular-path queries.
Therefore, \acp{ReGaP} solve a fundamentally different problem, thus the lack of comparison with the state of the art in regular-path queries.

\subsection{Impact of Node Merging}

\begin{table*}[t]
\centering
\begin{tabular}{l|rrrr|rrrr|}
\cline{2-9}
                                 & \multicolumn{4}{c|}{Base}                                                   & \multicolumn{4}{c|}{With Node Merging}                                                      \\ \cline{2-9} 
                                 & \multicolumn{1}{c|}{min} & \multicolumn{1}{c|}{max} & \multicolumn{1}{c|}{median} & average & \multicolumn{1}{c|}{min} & \multicolumn{1}{c|}{max} & \multicolumn{1}{c|}{median} & average \\ \hline
\multicolumn{1}{|l|}{\#Nodes}   & \multicolumn{1}{r|}{15}    & \multicolumn{1}{r|}{1 070}    & \multicolumn{1}{r|}{21}       &  26       & \multicolumn{1}{r|}{1}    & \multicolumn{1}{r|}{1 070}    & \multicolumn{1}{r|}{18}       &   22      \\ \hline
\multicolumn{1}{|l|}{\#Edges}   & \multicolumn{1}{r|}{14}    & \multicolumn{1}{r|}{1 335}    & \multicolumn{1}{r|}{24}       &   30.5      & \multicolumn{1}{r|}{0}    & \multicolumn{1}{r|}{1 335}    & \multicolumn{1}{r|}{21}       &  26      \\ \hline
\multicolumn{1}{|l|}{\#Var}    & \multicolumn{1}{r|}{285}    & \multicolumn{1}{r|}{31 601}    & \multicolumn{1}{r|}{1 745}       &        2 661 & \multicolumn{1}{r|}{5}    & \multicolumn{1}{r|}{25 259}    & \multicolumn{1}{r|}{1 423}       &   2 097      \\ \hline
\multicolumn{1}{|l|}{\#Const} & \multicolumn{1}{r|}{9675}    & \multicolumn{1}{r|}{21 079 098}    & \multicolumn{1}{r|}{228 175}       &       659 547  & \multicolumn{1}{r|}{15}    & \multicolumn{1}{r|}{17 343 652}    & \multicolumn{1}{r|}{164 140}       & 490 259        \\ \hline
\end{tabular}
\caption{Comparison of the graph and encoding size with and without node merging.}
\label{tb:node_merging_simplification}
\end{table*}

Table~\ref{tb:node_merging_simplification} shows the impact of node merging on the size of the graph and the \ac{SAT} encoding.
We only consider instances for which the \ac{ReGaP} matcher did not timeout before the encoding was complete, both with and without node merging.
Moreover, one of the \acp{ReGaP} used in this evaluation contains an edge between wildcards.
The respective instances are also not considered since node merging is not applicable in this scenario (see Proposition~\ref{prop:node_merging}).
The reduction in the number of nodes is 15.4\%, on average, which translates to a reduction, on average, of 25.7\% less clauses in the \ac{SAT} encoding.
We observed that the overhead of node merging is at most 1 second for these instances.

\begin{figure}
    \centering
    \includegraphics[width=\linewidth]{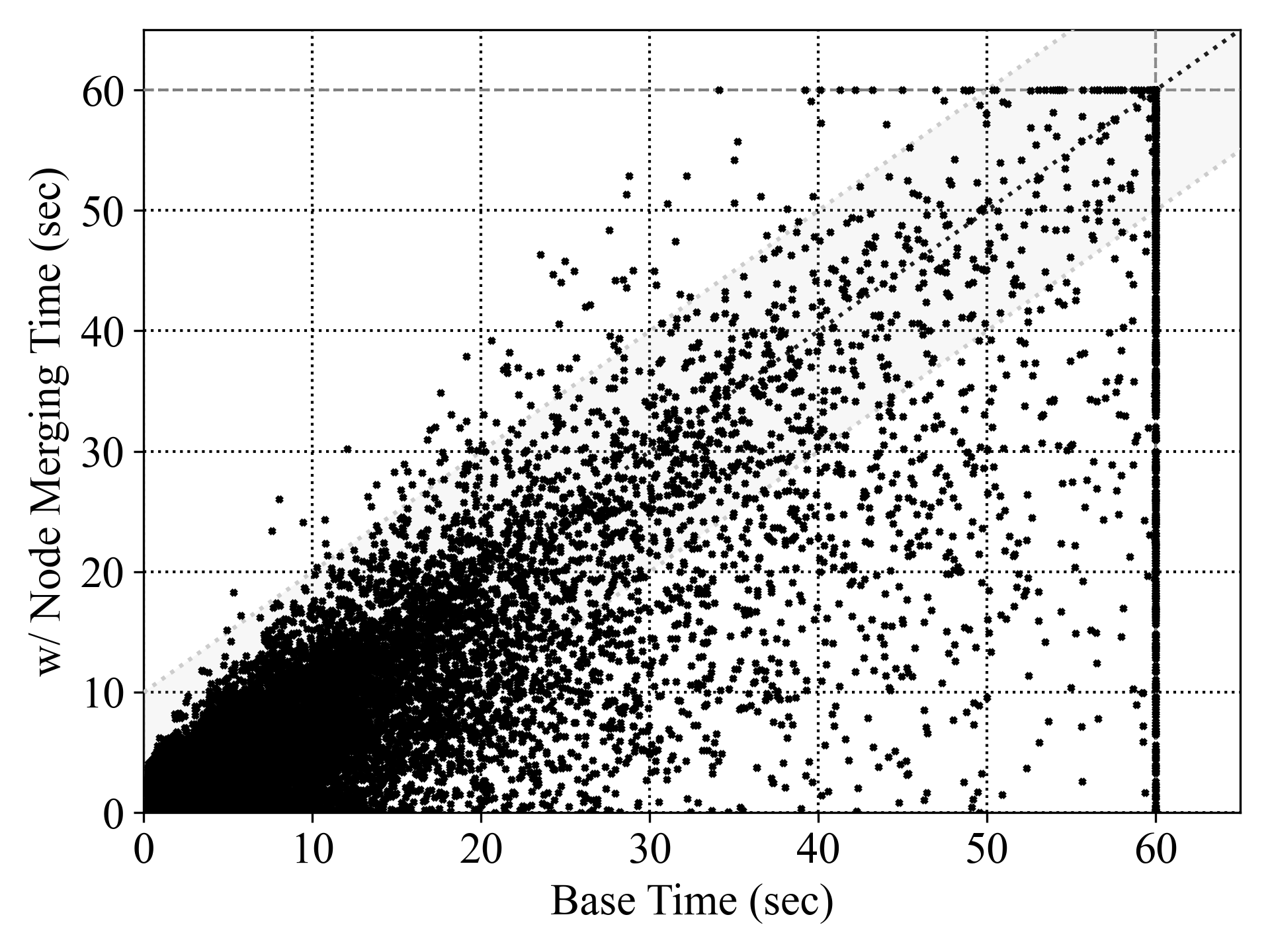}
    \caption{Execution time with node merging versus without.}
    \label{fig:config_time}
\end{figure}

Figure~\ref{fig:config_time} compares the execution time, in seconds, with and without node merging.
The \ac{ReGaP} with the edge between wildcards is also not considered in this figure.
Node merging has a significant impact on performance, being able to solve many more instances faster than the base encoding.
For example, 42 \nolinebreak 498 more instances are solved in less than 10 seconds with node merging.
971 of these instances result in a timeout without node merging.
The base encoding resulted in timeouts for 34 \nolinebreak 110 out of the 946 \nolinebreak 556 instances.
With node merging, this value is reduced to 26 \nolinebreak 487.
However, some instances are actually solved slower with node merging.
In fact, node merging times out for 940 instances that are solved with the base encoding.
We observed that node merging resulted in very little reduction for these specific instances: 0\% median reduction and only 1.8\% in the 90-percentile.
When it is 0\%, the time required by the base algorithm is very close to the timeout: around 58 seconds on average.
Therefore, these timeouts are likely due to noise introduced by worker contention in the parallel experimental environment.
When the reduction is very small, this causes the formula’s variables and constraints to change, which can trigger unpredictable behavior in the solver.
Because the size of the formula is very similar, the slightly smaller formula can be harder to solve.

\subsection{Impact of Graph Size}

\begin{figure}
    \centering
    \includegraphics[width=\linewidth]{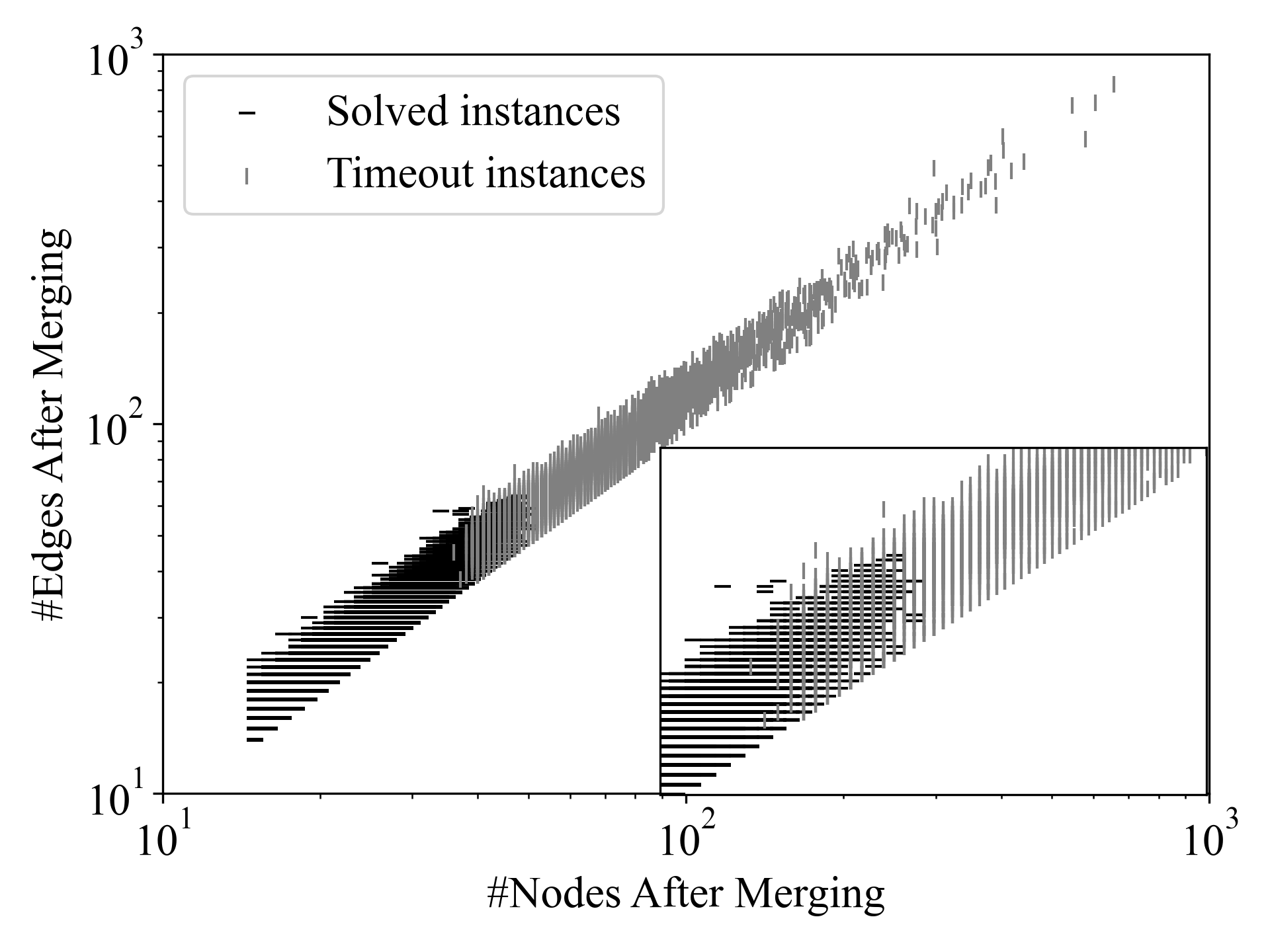}
    \caption{Timeouts as a function of graph size.}
    \label{fig:branched_termination_unconstrained-timeouts}
\end{figure}

Figure~\ref{fig:branched_termination_unconstrained-timeouts} shows the timeouts and solved instances as a function of graph size, considering only the \ac{ReGaP} with an edge between wildcards.
The maximum number of nodes (edges) among solved instances is 50 (64), while the minimum for the instances that timeout is 34 (38).
This hints at a strong potential for further performance improvements by investing in further simplification of the graph.
We observed the same behavior for the remaining \acp{ReGaP}, with some variation regarding the maximum graph size among solved instances and the minimum for the ones that resulted in timeouts.

\subsection{Impact of the Number of Wildcards}

\begin{table}[t]
\centering
\begin{tabular}{lcrr|rrr|}
\cline{5-7}
                                     & \multicolumn{1}{l}{}            & \multicolumn{1}{l}{} & \multicolumn{1}{l|}{} & \multicolumn{3}{c|}{\textbf{Exec Time (s)}}                                                          \\ \hline
\multicolumn{1}{|l|}{\textbf{ReGaP}} & \multicolumn{1}{c|}{\textbf{W}} & \multicolumn{2}{c|}{\textbf{Timeouts}}              & \multicolumn{1}{c|}{\textbf{avg}} & \multicolumn{1}{c|}{\textbf{med}} & \multicolumn{1}{c|}{\textbf{std}} \\ \hline
\multicolumn{1}{|l|}{call loops}     & \multicolumn{1}{c|}{1}          & 337                  & 0.5\%                 & \multicolumn{1}{r|}{1.3}          & \multicolumn{1}{r|}{0.2}          & 4.0                               \\ \hline
\multicolumn{1}{|l|}{call}           & \multicolumn{1}{c|}{2}          & 930                  & 1.3\%                 & \multicolumn{1}{r|}{2.7}          & \multicolumn{1}{r|}{0.6}          & 5.8                               \\ \hline
\multicolumn{1}{|l|}{ma loops}       & \multicolumn{1}{c|}{2}          & 1071                 & 1.5\%                 & \multicolumn{1}{r|}{3.1}          & \multicolumn{1}{r|}{0.8}          & 6.2                               \\ \hline
\multicolumn{1}{|l|}{afa loops s}    & \multicolumn{1}{c|}{3}          & 2003                 & 2.7\%                 & \multicolumn{1}{r|}{4.8}          & \multicolumn{1}{r|}{2.1}          & 7.9                               \\ \hline
\multicolumn{1}{|l|}{bterm loose}    & \multicolumn{1}{c|}{3}          & 8026                 & 11.0\%                & \multicolumn{1}{r|}{9.6}          & \multicolumn{1}{r|}{5.0}          & 11.5                              \\ \hline
\multicolumn{1}{|l|}{bterm}          & \multicolumn{1}{c|}{3}          & 2010                 & 2.8\%                 & \multicolumn{1}{r|}{4.5}          & \multicolumn{1}{r|}{1.8}          & 7.8                               \\ \hline
\multicolumn{1}{|l|}{fcall loops} & \multicolumn{1}{c|}{3}          & 1733                 & 2.4\%                 & \multicolumn{1}{r|}{4.5}          & \multicolumn{1}{r|}{1.8}          & 7.7                               \\ \hline
\multicolumn{1}{|l|}{ma}             & \multicolumn{1}{c|}{3}          & 1785                 & 2.4\%                 & \multicolumn{1}{r|}{4.5}          & \multicolumn{1}{r|}{1.8}          & 7.8                               \\ \hline
\multicolumn{1}{|l|}{afa loops}      & \multicolumn{1}{c|}{4}          & 2915                 & 4.0\%                 & \multicolumn{1}{r|}{6.4}          & \multicolumn{1}{r|}{3.2}          & 9.1                               \\ \hline
\multicolumn{1}{|l|}{fcall}       & \multicolumn{1}{c|}{4}          & 2737                 & 3.8\%                 & \multicolumn{1}{r|}{6.2}          & \multicolumn{1}{r|}{3.1}          & 9.0                               \\ \hline
\multicolumn{1}{|l|}{mfy loops}      & \multicolumn{1}{c|}{4}          & 2757                 & 3.8\%                 & \multicolumn{1}{r|}{6.4}          & \multicolumn{1}{r|}{3.2}          & 9.1                               \\ \hline
\multicolumn{1}{|l|}{afa}            & \multicolumn{1}{c|}{5}          & 3983                 & 5.5\%                 & \multicolumn{1}{r|}{8.0}          & \multicolumn{1}{r|}{4.2}          & 10.1                              \\ \hline
\multicolumn{1}{|l|}{mfy}            & \multicolumn{1}{c|}{5}          & 3823                 & 5.2\%                 & \multicolumn{1}{r|}{7.9}          & \multicolumn{1}{r|}{4.2}          & 10.1                              \\ \hline
\end{tabular}
\caption{The number of wildcards (W column) and timeouts, and the baseline execution times for each \ac{ReGaP}.}
\label{tab:regap_time}
\end{table}

Table~\ref{tab:regap_time} compares the number of timeouts and execution times obtained without node merging for each \ac{ReGaP}.
The rows are sorted by the respective number of wildcards.
Note that the execution time statistics do not consider instances that resulted in a timeout.
In most cases, the number of timeouts and the mean/median execution time seem to grow with the number of wildcards.
The algorithm also seems to become more unstable as the number of wildcards grows, as evidenced by the increase in the standard deviation.

The obvious exception is the \textit{bterm loose} \ac{ReGaP}, the one with edges between wildcard nodes, which was expected for the following reason.
Lets consider an edge $(w, w')$ between, for example, two any-1+-subgraph wildcards $w$ and $w'$.
The expansion of $w$ replaces the edge $(w, w')$ by $\left| V \right|$ new edges from each of the non-wildcard nodes inserted in place of $w$ to $w'$.
In turn, each of those edges will be replaced by another $\left| V \right|$ edges when $w'$ is expanded as well.

Note that this is a preliminary evaluation and further experiments should be performed with a more diverse set of \acp{ReGaP}.
Also, this pattern of performance degradation as the number of wildcards increases is not so clear with node merging.
Depending on the narrowness of the constraints, node merging can have a significant impact on performance. 

\section{Conclusions \& Future Work}
\label{sec:conclusion}

Solving graph matching problems has many applications.
In particular, it is an essential tool for code analysis in visual programming languages.
However, the state of the art focuses either on solving graph isomorphism, approximated graph matching or regular-path queries.
We propose \ac{ReGaP} matching, an extension of graph isomorphism that allows one to check complex structural properties through declarative specifications.
We propose a \ac{SAT} encoding for solving \ac{ReGaP} matching and a simplification technique for reducing encoding size, thus improving the performance of the \ac{SAT} solver.
An extensive experimental evaluation carried on benchmarks from the CodeSearchNet dataset~\cite{DBLP:journals/corr/abs-1909-09436} shows the effectiveness of the proposed approach.

In the future, we plan to extend \acp{ReGaP} with new types of wildcards (e.g. optional nodes/edges and sequence/subgraph wildcards with size limitations).
We also wish to explore more compact encodings for \ac{ReGaP} matching that do not rely on wildcard expansion, as well as further evaluate the impact of the number of wildcards and overall structure of the \acp{ReGaP} on the performance of the algorithm.
Other \ac{SAT} solvers and alternative automated reasoning frameworks, such as constraint programming~\cite{DBLP:reference/fai/2} and satisfiability modulo theories~\cite{DBLP:journals/cacm/MouraB11}, should also be evaluated.
We can also explore tighter bounds for the value of $k$ used for expansion, and an algorithm that starts with a small value for $k$ and iteratively increments $k$ until the formula becomes satisfiable or an upper bound is reached.
Lastly, we also plan to develop tools for synthesizing \acp{ReGaP} from positive and negative examples.
Note that \ac{ReGaP} matching is a general problem that we believe has the potential to be useful in other applications that deal with graph data, such as computational biology, chemistry and network analysis.
We hope to see future work explore such applications.

\bibliography{aaai24}

\end{document}